\documentclass{article}
\usepackage{jmlr2e,amsmath,graphicx}
\usepackage{times}
\usepackage{epsfig}
\usepackage{amssymb} 
\usepackage{color}
\usepackage[]{algorithm2e}
\usepackage{newfloat}
\usepackage{listings, bbding,pifont}
\usepackage{algpseudocode}
\usepackage{xspace}
\usepackage{url}
\usepackage{mathtools}
\usepackage{booktabs}
\usepackage{hhline}
\usepackage{wrapfig}
\usepackage{cleveref}
\usepackage[normalem]{ulem}
\useunder{\uline}{\ul}{}
\usepackage[table,xcdraw]{xcolor}
 \usepackage{multirow}

\newcommand{\alg}[0]{\texttt{SiSTA}}
\newcommand{\jay}[1]{\textcolor{orange}{[Jay]}}
\newcommand{\rak}[1]{\textcolor{green}{[Rak]}}
\newcommand{\kow}[1]{\textcolor{red}{[Kow]}}

\begin{document}
\title{Single-Shot Domain Adaptation via Target-Aware Generative Augmentations}

\author{\name Rakshith Subramanyam \email  rakshith.subramanyam@asu.edu \\
       \addr Arizona State University\\
       Tempe, AZ, USA.
       \AND
       \name Kowshik Thopalli \email kthopall@asu.edu \\
       \addr Arizona State University\\
       Tempe, AZ, USA.
       \AND
       \name Spring Berman \email spring.berman@asu.edu \\
       \addr Arizona State University\\
       Tempe, AZ, USA.
       \AND
       \name Pavan Turaga  \email pturaga@asu.edu\\
       \addr Arizona State University \\
       Tempe, AZ, USA.
       \AND
       \name Jayaraman J. Thiagarajan \email jjayaram@llnl.gov \\
       \addr Lawrence Livermore National Laboratory \\
       Livermore, CA, USA.}

\editor{}

\maketitle

\begin{abstract}
The problem of adapting models from a source domain using data from any target domain of interest has gained prominence, thanks to the brittle generalization in deep neural networks. While several test-time adaptation techniques have emerged, they typically rely on synthetic data augmentations in cases of limited target data availability. In this paper, we consider the challenging setting of single-shot adaptation and explore the design of augmentation strategies. We argue that augmentations utilized by existing methods are insufficient to handle large distribution shifts, and hence propose a new approach \alg~(\underline{Si}ngle-\underline{S}hot \underline{T}arget \underline{A}ugmentations), which first fine-tunes a generative model from the source domain using a single-shot target, and then employs novel sampling strategies for curating synthetic target data. Using experiments with a state-of-the-art domain adaptation method, we find that \alg~produces improvements as high as $20\%$ over existing baselines under challenging shifts in face attribute detection, and that it performs competitively to oracle models obtained by training on a larger target dataset.

\end{abstract}

\section{Introduction}
\label{sec:intro}

Despite producing high accuracies in the \textit{i.i.d.} setting, deep models are known to fail unpredictably under real-world distribution shifts (or domain shifts)~\citep{torralba2011unbiased}. Such failures can be potentially mitigated by refining the model weights with data from the target domain of interest. A large class of approaches have been explored in this regard; popular examples include source free domain adaptation (SFDA)~\citep{SHOT} and test-time adaptation (TTA)~\citep{Tent}. Not surprisingly, the effectiveness of these approaches can be significantly limited when sufficient target data is not available. In this paper, we consider the extreme scenario where only single-shot target data is accessible. 

Driven by the data scarcity challenge in practical settings, data augmentation has emerged as a common fix for enabling model adaptation even with limited data For example, the recently proposed MEMO~\citep{zhang2021memo} leverages pre-specified image augmentations (e.g., Augmix~\citep{hendrycks2019augmix}) to expand the limited target data and performs test-time adaptation. Note, the success of such approaches directly hinges on how well the chosen augmentation can represent the target data distribution, and hence, in practice, different augmentation techniques may lead to varying degrees of generalization.

\begin{figure*}[t]
    \centering
    \includegraphics[width=1\textwidth]{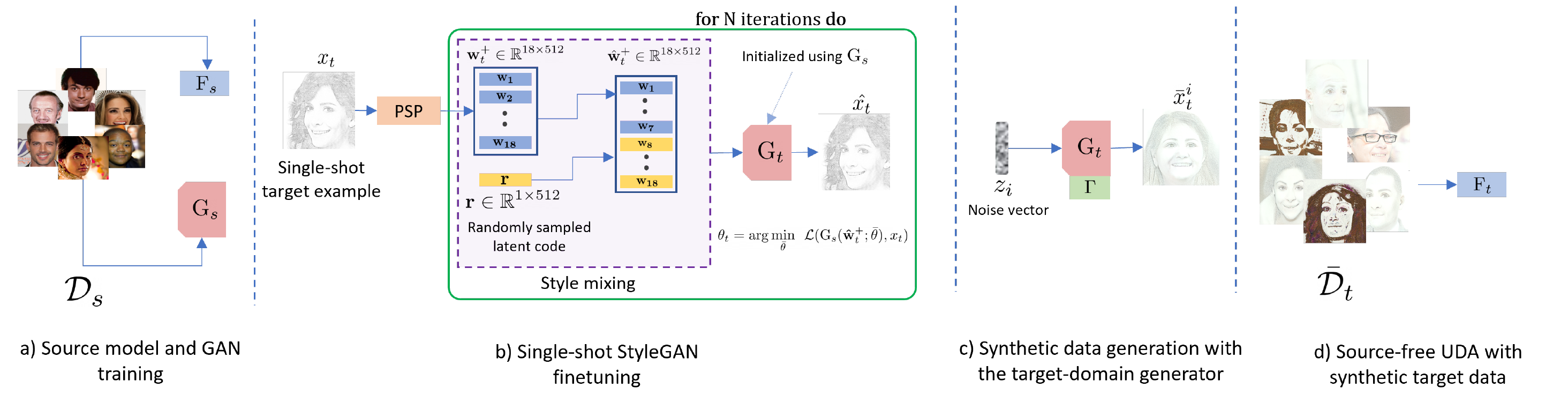}
    \caption{\alg: Assuming access to both the classifier and a StyleGAN from the source domain, we first adapt the generator to the target domain using a single-shot example. Next, we employ the proposed activation pruning strategies to construct the synthetic target dataset $\bar{\mathcal{D}}_t$. Finally, this dataset is used with any test-time adaptation technique for model refinement.}
    \label{fig:pipeline}
\end{figure*}

With the goal of advancing test-time adaptation with single-shot target data, we propose \alg~a target domain-aware augmentation technique to synthetically generate target data, which can be used with any unsupervised domain adaption method for improving model generalization. At its core, our method relies on deep generative models, in particular StyleGANv2~\citep{karras2019style}, for data synthesis. To this end, \alg~first adapts the source StyleGAN using a training strategy inspired from~\citep{chong2021jojogan}, and subsequently employs novel activation pruning strategies for sampling the target StyleGAN and curating a synthetic target dataset. Finally, this unlabeled dataset is used in conjunction with any SFDA method~\citep{yang2021exploiting} to adapt source classifiers. Using empirical studies with multiple face attribution detection tasks and a variety of distribution shifts, we show that \alg~significantly outperforms existing approaches and that it performs competitively to \textit{oracle} models obtained by adapting with large target domain datasets.

\section{Background}
\label{sec:background}
Data augmentation has become an important tool to develop generalizable models, especially when operating in limited data settings.
It has been shown that data augmentation can improve both in-distribution and out-of-distribution (OOD) accuracies~\citep{steiner2021train,hendrycks2021many}. 
Existing augmentations can be broadly viewed in two categories - (i) pixel/geometric corruptions and (ii) generative augmentations. The former category includes strategies such as CutMix~\citep{cutmix2}, Cutout~\citep{cutout1}, Augmix~\citep{hendrycks2019augmix}, RandConv~\citep{randconv}, mixup~\citep{mixup} and AutoAugment~\citep{autoaugment1}. These domain-agnostic methods are known to be insufficient to achieve OOD generalization, especially under large domain shifts. To circumvent this, more recent solutions have resorted to generative models (e.g., GANs) for synthesizing plausible augmentations~\citep{SurveyGanAug}. Specifically, popular methods such as MBDG~\citep{robey2021model}, CyCADA~\citep{hoffman2018cycada} and GenToAdapt~\citep{sankaranarayanan2018generate} have leveraged generative augmentations to better adapt to unlabeled target domains. However, these methods can be ineffective in cases of limited target data availability. In this work, we consider the extreme setting of single-shot target data and assume no access to source data during adaptation. Our goal is to obtain generative augmentations using only single-shot target, which can then be used in conjunction with any existing SFDA technique~\citep{yang2021exploiting, SHOT, yeh2021sofa, Tent}.

\section{Proposed Approach}
\label{sec:approach}


We investigate the problem of adapting source domain classifiers using a single-shot target example and propose \alg, a target domain-aware augmentation technique (see Figure \ref{fig:pipeline}).


\noindent \textbf{Setup.} Formally, we denote the labeled source data as $\mathcal{D}_s = \{(x_s^i, y_s^i)\}$ with images $x_s^i$ and labels $y_s^i$ and the single-shot target example as $x_t$. We assume that we have access to both the classifier $\mathrm{F}_s: x \rightarrow y$ with parameters $\Phi_s$ and the StyleGAN-v2 model (generator $\mathrm{G}_s: z \rightarrow x$ with parameters $\Theta_s$ and discriminator $\mathrm{H}_s$) trained on the source dataset $\mathcal{D}_s$. Our goal is to generate a synthetic target dataset $\bar{\mathcal{D}}_t = \{\bar{x}_t^i\}$ and refine the source classifier to obtain the target hypothesis $\mathrm{F}_t(.;\Phi_t)$ using any domain adaptation method.

\noindent \textbf{Step 1: Source training.} We begin by training the source classifier $\mathrm{F}_s$ using labeled data $\mathcal{D}_s$. This is carried out using the cross entropy loss and standard training configurations. In addition, we build a generative model for the source data distribution. More specifically, we use the StyleGAN-v2 architecture and infer $\mathrm{G}_s$ and $\mathrm{H}_s$ respectively.

\noindent \textbf{Step 2: Single-shot StyleGAN finetuning.} Next, we fine-tune $\mathrm{G}_s$ using only the single-shot example $x_t$, in order to generate images from the target domain. To this end, we first invert $x_t$ onto the style space of $\mathrm{G}_s$ using a pre-trained encoder, e.g., Pixel2Style2Pixel or shortly PSP~\citep{richardson2021encoding}, which maps a given image into the style code $\mathbf{w}_t^+ \in \mathbb{R}^{18\times512}$. This latent code corresponds to $18$ intermediate layers of StyleGAN-v2. By design, $x_t$ may be out of the training distribution $P_s(x)$ and hence the reconstruction corresponding to $\mathbf{w}_t^+$ is more likely to resemble the source domain. Consequently, we need to refine the generator $\mathrm{G}_s$ to synthesize images that are characteristic of the target domain.

\RestyleAlgo{boxruled}
    \begin{center}
    \begin{algorithm}[t]
    \small
	\KwIn  {Target sample $x_t$,
	        No. of training iterations $N$,
	        Source generator $\mathrm{G}_s$, 
	        PSP encoder $\mathrm{E}$.
	        }
    \KwOut{
    Target domain StyleGAN $\mathrm{G}_t$.
    } 
    Invert the target sample to obtain  $\mathbf{w}_t^+ = \mathrm{E}(x_t)$\;
        
    \For {$n$ in $1$ to $N$}
    {
        Generate random style latent $\mathbf{r}\in \mathbb{R}^{1 \times 512}$\;
        Perform style-mixing, \textit{i.e.}, replace layers $8$-$18$ of $\mathbf{w}^+_t$ with $\mathbf{r}$\;
        Generate image $\hat{x}_t = \mathrm{G}_s(\mathbf{\hat{w}}^+_t)$\;
        Update parameters $\Theta_t = \arg \min_{\bar{\Theta}} \mathcal{L}(\hat{x}_t, x_t; \mathrm{H}_s)$\;
    }
    \Return $\mathrm{G}_t$ with parameters $\Theta_t$.
	\caption{Single-shot StyleGAN fine-tuning}
	\label{algo1}
\end{algorithm}
\vspace{-0.35in}
\end{center}

We take inspiration from JoJoGAN~\citep{chong2021jojogan}, a recent optimization strategy for style transfer in GANs, and update the generator model parameters based on a loss function defined on the activation outputs from the frozen discriminator $\mathrm{H}_s$: 
\begin{align}
   \Theta_t = \arg \min_{\bar{\Theta}} \phantom{r} \sum_{\ell} \|\mathrm{H}^{\ell}_s(\mathrm{G}_s(\mathbf{w}_t^+;\bar{\Theta})) - \mathrm{H}_s^{\ell}(x_t)\|,
    \label{eq:ganInversion}
\end{align}where $\Theta_t$ refers to the parameters of the updated generator $\mathrm{G}_t$, $\mathrm{H}^{\ell}_s$ denotes the activations from layer $\ell$ of the discriminator $\mathrm{H}_s$, and this objective minimizes the discrepancy between the target image and the reconstruction from the generator. Since this optimization can be highly unstable with a single $x_t$, we construct attribute-shifted versions of $x_t$ through a style-mixing protocol, wherein the latent codes corresponding to a pre-specified subset of layers in $\mathbf{w}_t^+$ are replaced with randomly generated latents obtained by transforming a noise vector $z \sim \mathcal{N}(\mathbf{0, I})$ with the mapping network in StyleGAN-v2. In particular, we replace the layers $8$ to $18$ of $\mathbf{w}_t^+$, as it is known that the initial layers encode the key semantic content, while the later layers contain style characteristics. In each iteration of our optimization, a different style-mixed latent code $\mathbf{\hat{w}}_t^+$ is used with \eqref{eq:ganInversion}. Algorithm \ref{algo1} lists this procedure.

\RestyleAlgo{boxruled}
    \begin{center}
    \begin{algorithm}[t]
    \small
	\KwIn{Target GAN $\mathrm{G}_t(.;\Theta_t)$, Source GAN $\mathrm{G}_s(.;\Theta_s)$,
	Pruning strategy $\mathrm{\Gamma}$,
	Pruning ratio $p$}
    \KwOut{Sampled image $\bar{x}_t$ } 
    Generate a random style latent $\mathbf{w}^+ \in \mathbb{R}^{18 \times 512}$\;
    
    \For{$\ell$ in $8$ to $18$}
    {
        $s \sim \text{RandInt}(0,1)$\;
        \If{$s==1$}
        {   
            Obtain layer $\ell$ activations $h_t^{\ell}$ from $\mathrm{G}_t(\mathbf{w}^+)$\;
            \For{$k$ in $1$ to $K^{\ell}$}
            {   
                $\tau_p = p$-th percentile of $h_t^{\ell}[:,:,k]$\;
                \eIf{$\mathrm{\Gamma}==$ \color{blue}{\textit{prune-zero}}}
                {
                    $h_t^{\ell}[i,j,k] = 0$ \textbf{if} $h_t^{\ell}[i,j,k] < \tau_p, \forall i, j$\;
                }
                {
                    Obtain activations $h_s^{\ell}$ from $\mathrm{G}_s(\mathbf{w}^+)$\;
                    $h_t^{\ell}[i,j,k] = h_s^{\ell}[i,j,k]$ \textbf{if} $h_t^{\ell}[i,j,k] < \tau_p, \forall i, j$\;
                }
            }
        } 
    }    
    \Return Image $\bar{x}_t = \mathrm{G}_t(\mathbf{w}^+; \mathrm{\Gamma})$
	\caption{Generating synthetic target data}
	\label{algo2}
\end{algorithm}
\vspace{-0.45in}
\end{center}

\begin{figure}[t]
    \centering
    \includegraphics[width=0.99\columnwidth]{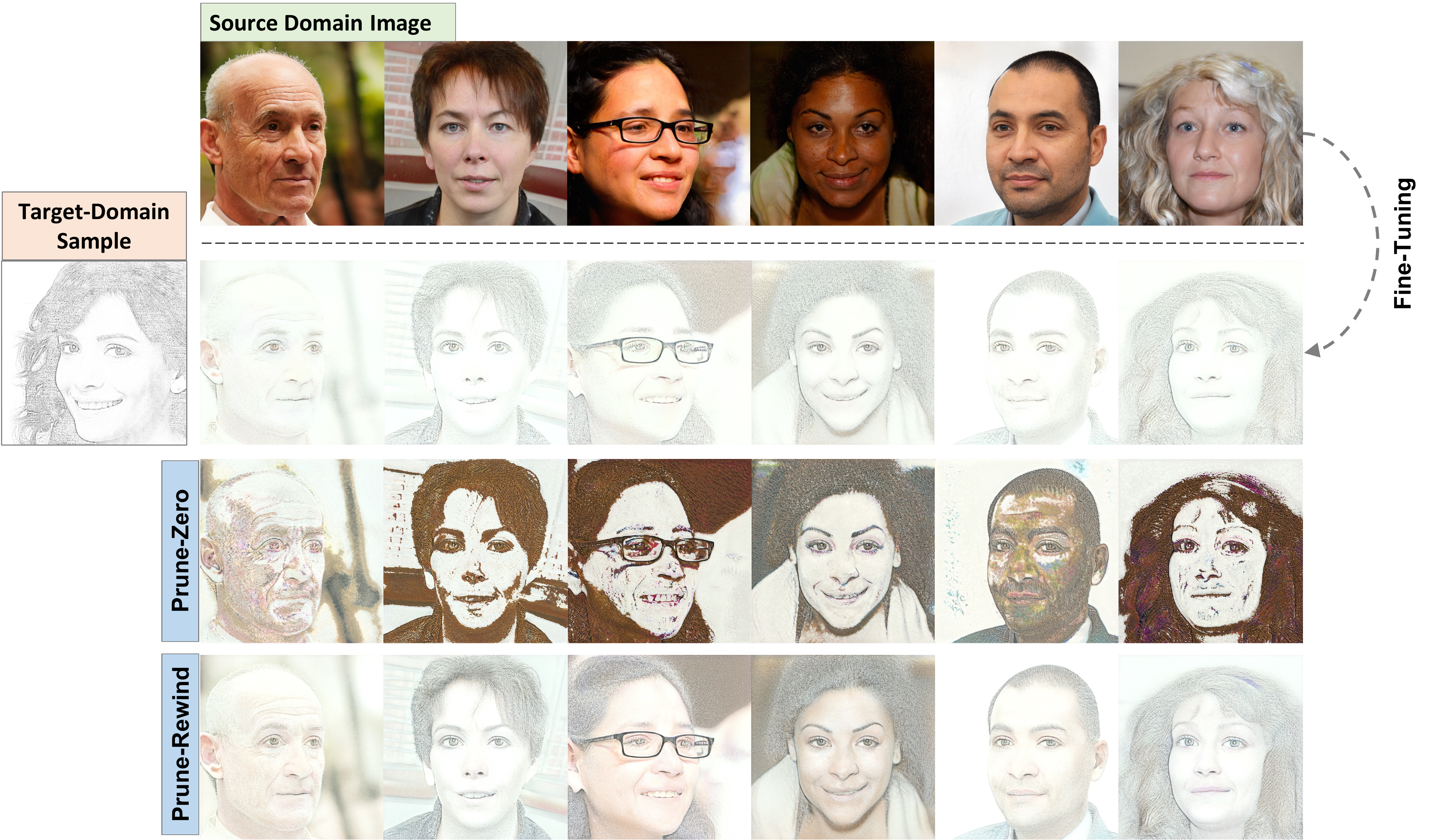}
    \caption{Synthetic data generated using our proposed approach. In each case, we show the source domain image and the corresponding reconstructions from the target StyleGAN sampling (base), prune-zero and prune-rewind strategies.}
    \label{fig:front_page}
\end{figure}

\noindent \textbf{Step 3: Synthetic data generation.}
Once we obtain the adapted StyleGAN generator $\mathrm{G}_t$ for the target domain, we can build our synthetic dataset by sampling in its latent space. Despite the efficacy of such an approach, the inherent discrepancy between the true target distribution $P_t(x)$ and the approximate $Q_t(x)$ (synthetic data) can limit the generalization. We propose to address this by perturbing the latent representations from different layers of $\mathrm{G}_t$ to realize a more diverse set of style variations. More specifically, we introduce two strategies based on activation pruning, which identify all activations (at the output of each layer) that are lower than the $p^\text{th}$ percentile value and replace them with zero (referred as \textit{prune-zero}) or with corresponding activations from the source GAN $\mathrm{G}_s$ (\textit{prune-rewind}). While the former strategy attenuates the effect of the target generator neurons to synthesize variations, the latter attempts to implicitly sample along the geodesic between the source and target domains by mixing the activations from the two generators. Note, we perform the pruning only in layers $8$-$18$ so that the semantic content of a sample is not changed. Algorithm \ref{algo2} describes the sampling process and Figure \ref{fig:front_page} illustrates the synthetic data generated for a target domain (\textit{pencil sketch}) using vanilla sampling (or base), prune-zero and prune-rewind strategies.

\noindent \textbf{Step 4: Source-free unsupervised domain adaptation:}
Using the synthetically generated target domain data, we finally perform source-free adaptation of $\mathrm{F}_s$ to obtain the target hypothesis $\mathrm{F}_t$. To this end, we employ NRC~\citep{yang2021exploiting}, a state-of-the-art SFDA method\footnote{\url{https://github.com/Albert0147/NRC_SFDA}}, which exploits the intrinsic neighborhood structure of the target data and pushes samples close to their semantically close neighbors by enforcing prediction consistency. We refer to~\citep{yang2021exploiting} for additional details.

\begin{figure}[t]
\centering
    \includegraphics[width=0.35\linewidth]{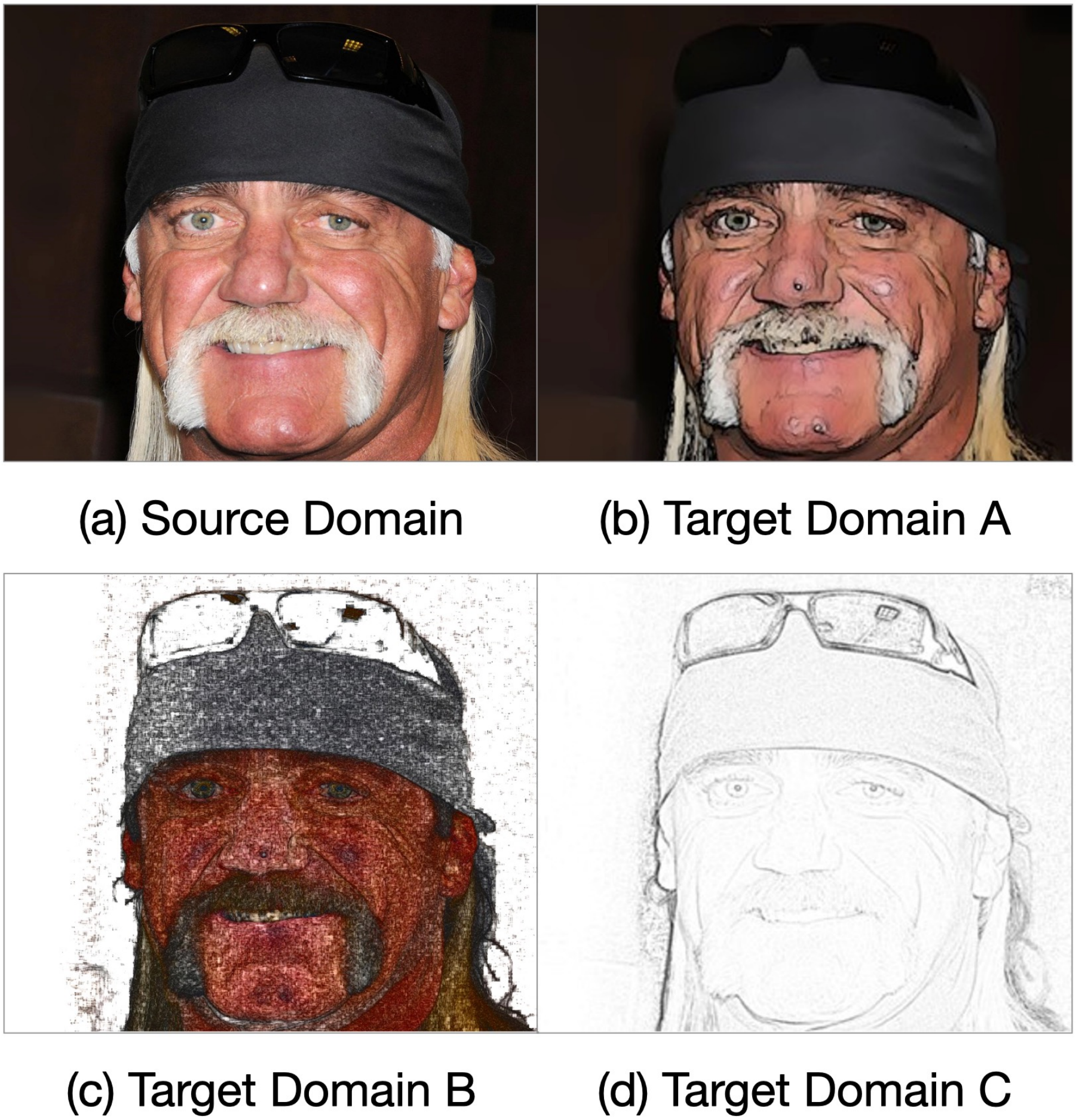}
    \caption{We emulate three real-world distribution shifts with increasing levels of severity.}
    \label{fig:domains}
    \vspace{-0.1in}
\end{figure}


\section{Empirical Results}
\label{sec:results}
\noindent \textbf{Dataset.} For our empirical study, we consider the task of face attribute detection with images from the CelebA-HQ dataset. We used the pre-trained StyleGAN-v2 from~\citep{karras2019style} and emulated three different distribution shifts (referred as domains A, B, C in Figure \ref{fig:domains}). CelebA-HQ is a high-quality large-scale face attribute dataset with 30000 images, which is split into a source dataset with 18K images and the rest was used to construct the target domains. To emulate varying levels of distribution shift, we employed standard image manipulation techniques (we release this new benchmark dataset along with our codes\footnote{\textbf{\alg}: \url{https://github.com/kowshikthopalli/SISTA}}): (i) \textit{Domain A}: We used the Stylization technique in OpenCV with $\sigma_s=40$ and $\sigma_r = 0.2$; (ii) \textit{Domain B}: For this shift, we used the PencilSketch technique in OpenCV with $\sigma_s=40$ and $\sigma_r = 0.04$; and (iii) \textit{Domain C}: This challenging domain shift was created by converting each color image to grayscale, and then performing pixel-wise division with a smoothed, inverted grayscale image. In our experiments, one randomly chosen example from each target domain was used for performing adaptation, and the performance on the entire target set of $12,000$ images is reported. 
We consider $4$ facial attribute detection tasks: (i) \textit{Smiling} (ii) \textit{Gender} (iii) \textit{Arched Eyebrows} and (iv) \textit{Mouth Slightly Open}.  

\noindent \textbf{Experiment Setup.} 

\noindent (a) \underline{\textit{Source model training}}: To obtain the source model $\mathrm{F}_s$ we fine-tune a Imagenet pre-trained ResNet-$50$~\citep{he2016deep} with labeled source data. We use a learning rate of $1e-4$, Adam optimizer and train for $30$ epochs; 

\noindent (b) \underline{\textit{StyleGAN fine-tuning}}: For Algorithm \ref{algo1}, we set $N = 300$; 

\noindent (c) \underline{\textit{Synthetic data curation}}: In Algorithm \ref{algo2}, we set $p = 20\%$ for prune-rewind and $p = 50\%$ for prune-zero strategies, and generated $1000$ samples in each case. Note, we experimented by varying the size of $\bar{\mathcal{D}}_t$ (between $100$ and $10,000$). We found the performance to improve steadily until $1000$ and no significant benefits were observed beyond $1000$.; 

\noindent (d) \underline{\textit{NRC SFDA training}}: For NRC, we set both neighborhood and expanded neighborhood sizes at $5$. Finally, we adapt $\mathrm{F}_s$ using SGD with momentum of 0.9 and learning rate of $1e-3$.

\begin{table*}[t]
\label{tab:results}
\renewcommand{\arraystretch}{1.3}
\small
\centering
\resizebox{0.8\linewidth}{!}{
\begin{tabular}{clcccclcccc}

\cline{1-1} \cline{3-6} \cline{8-11}
\multicolumn{1}{|c|}{\cellcolor[HTML]{00009B}{\color[HTML]{FFFFFF} }}                                   & \multicolumn{1}{l|}{\cellcolor[HTML]{FFFFFF}}                   & \multicolumn{4}{c|}{\cellcolor[HTML]{00009B}{\color[HTML]{FFFFFF} \textbf{Attribute: Smiling}}}                                                                                                                                                                                                                                                   & \multicolumn{1}{l|}{\cellcolor[HTML]{FFFFFF}}                   & \multicolumn{4}{c|}{\cellcolor[HTML]{00009B}{\color[HTML]{FFFFFF} \textbf{Attribute: Gender}}}                                                                                                                                                                          \\ \cline{3-6} \cline{8-11} 
\multicolumn{1}{|c|}{\multirow{-2}{*}{\cellcolor[HTML]{00009B}{\color[HTML]{FFFFFF} \textbf{Methods}}}} & \multicolumn{1}{l|}{\cellcolor[HTML]{FFFFFF}}                   & \multicolumn{1}{c|}{\cellcolor[HTML]{DAE8FC}\textbf{Domain A}}                     & \multicolumn{1}{c|}{\cellcolor[HTML]{DAE8FC}\textbf{Domain B}}                     & \multicolumn{1}{c|}{\cellcolor[HTML]{DAE8FC}\textbf{Domain C}}                     & \multicolumn{1}{c|}{\cellcolor[HTML]{DAE8FC}\textbf{Average}}                      & \multicolumn{1}{l|}{\cellcolor[HTML]{FFFFFF}}                   & \multicolumn{1}{c|}{\cellcolor[HTML]{DAE8FC}\textbf{Domain A}} & \multicolumn{1}{c|}{\cellcolor[HTML]{DAE8FC}\textbf{Domain B}} & \multicolumn{1}{c|}{\cellcolor[HTML]{DAE8FC}\textbf{Domain C}}    & \multicolumn{1}{c|}{\cellcolor[HTML]{DAE8FC}\textbf{Average}}     \\ \cline{1-1} \cline{3-6} \cline{8-11} 
\rowcolor[HTML]{FFFFFF} 
\multicolumn{1}{|c|}{\cellcolor[HTML]{FFFFFF}{\color[HTML]{000000} Source only}}                        & \multicolumn{1}{l|}{\cellcolor[HTML]{FFFFFF}}                   & \multicolumn{1}{c|}{\cellcolor[HTML]{FFFFFF}{\color[HTML]{000000} \textbf{89.78}}} & \multicolumn{1}{c|}{\cellcolor[HTML]{FFFFFF}{\color[HTML]{000000} 73.68}}          & \multicolumn{1}{c|}{\cellcolor[HTML]{FFFFFF}{\color[HTML]{000000} 62.18}}          & \multicolumn{1}{c|}{\cellcolor[HTML]{FFFFFF}{\color[HTML]{000000} 75.21}}          & \multicolumn{1}{l|}{\cellcolor[HTML]{FFFFFF}}                   & \multicolumn{1}{c|}{\cellcolor[HTML]{FFFFFF}94.47}             & \multicolumn{1}{c|}{\cellcolor[HTML]{FFFFFF}83.72}             & \multicolumn{1}{c|}{\cellcolor[HTML]{FFFFFF}69.43}                & \multicolumn{1}{c|}{\cellcolor[HTML]{FFFFFF}82.54}                \\ \cline{1-1} \cline{3-6} \cline{8-11} 
\rowcolor[HTML]{FFFFFF} 
\multicolumn{1}{|c|}{\cellcolor[HTML]{FFFFFF}{\color[HTML]{000000} MEMO (AugMix)}}                      & \multicolumn{1}{l|}{\cellcolor[HTML]{FFFFFF}}                   & \multicolumn{1}{c|}{\cellcolor[HTML]{FFFFFF}{\color[HTML]{000000} \ul{89.34}}}          & \multicolumn{1}{c|}{\cellcolor[HTML]{FFFFFF}{\color[HTML]{000000} 71.76}}          & \multicolumn{1}{c|}{\cellcolor[HTML]{FFFFFF}{\color[HTML]{000000} 59.43}}          & \multicolumn{1}{c|}{\cellcolor[HTML]{FFFFFF}{\color[HTML]{000000} 73.51}}          & \multicolumn{1}{l|}{\cellcolor[HTML]{FFFFFF}}                   & \multicolumn{1}{c|}{\cellcolor[HTML]{FFFFFF}94.04}             & \multicolumn{1}{c|}{\cellcolor[HTML]{FFFFFF}82.45}             & \multicolumn{1}{c|}{\cellcolor[HTML]{FFFFFF}58.51}                & \multicolumn{1}{c|}{\cellcolor[HTML]{FFFFFF}78.33}                \\ \cline{1-1} \cline{3-6} \cline{8-11} 
\rowcolor[HTML]{FFFFFF} 
\multicolumn{1}{|c|}{\cellcolor[HTML]{FFFFFF}{\color[HTML]{000000} MEMO (RandConv)}}                    & \multicolumn{1}{l|}{\cellcolor[HTML]{FFFFFF}}                   & \multicolumn{1}{c|}{\cellcolor[HTML]{FFFFFF}{\color[HTML]{000000} {\ul 89.37}}}    & \multicolumn{1}{c|}{\cellcolor[HTML]{FFFFFF}{\color[HTML]{000000} 71.80}}          & \multicolumn{1}{c|}{\cellcolor[HTML]{FFFFFF}{\color[HTML]{000000} 59.27}}          & \multicolumn{1}{c|}{\cellcolor[HTML]{FFFFFF}{\color[HTML]{000000} 73.48}}          & \multicolumn{1}{l|}{\cellcolor[HTML]{FFFFFF}}                   & \multicolumn{1}{c|}{\cellcolor[HTML]{FFFFFF}94.05}             & \multicolumn{1}{c|}{\cellcolor[HTML]{FFFFFF}82.45}             & \multicolumn{1}{c|}{\cellcolor[HTML]{FFFFFF}58.76}                & \multicolumn{1}{c|}{\cellcolor[HTML]{FFFFFF}78.42}                \\ \cline{1-1} \cline{3-6} \cline{8-11} 
\rowcolor[HTML]{FFFFFF} 
\multicolumn{1}{|c|}{\cellcolor[HTML]{FFFFFF}{\color[HTML]{000000} Ours (base)}}                        & \multicolumn{1}{l|}{\cellcolor[HTML]{FFFFFF}}                   & \multicolumn{1}{c|}{\cellcolor[HTML]{FFFFFF}{\color[HTML]{000000} 84.80}}          & \multicolumn{1}{c|}{\cellcolor[HTML]{FFFFFF}{\color[HTML]{000000} 82.53}}          & \multicolumn{1}{c|}{\cellcolor[HTML]{FFFFFF}{\color[HTML]{000000} {83.29}}} & \multicolumn{1}{c|}{\cellcolor[HTML]{FFFFFF}{\color[HTML]{000000} 83.54}}          & \multicolumn{1}{l|}{\cellcolor[HTML]{FFFFFF}}                   & \multicolumn{1}{c|}{\cellcolor[HTML]{FFFFFF}\textbf{94.73}}    & \multicolumn{1}{c|}{\cellcolor[HTML]{FFFFFF}{\ul 87.52}}       & \multicolumn{1}{c|}{\cellcolor[HTML]{FFFFFF}89.44}                & \multicolumn{1}{c|}{\cellcolor[HTML]{FFFFFF}90.56}                \\ \cline{1-1} \cline{3-6} \cline{8-11} 
\rowcolor[HTML]{FFFFFF} 
\multicolumn{1}{|c|}{\cellcolor[HTML]{FFFFFF}{\color[HTML]{000000} Ours (prune-zero)}}                & \multicolumn{1}{l|}{\cellcolor[HTML]{FFFFFF}}                   & \multicolumn{1}{c|}{\cellcolor[HTML]{FFFFFF}{\color[HTML]{000000} {88.65}}} & \multicolumn{1}{c|}{\cellcolor[HTML]{FFFFFF}{\color[HTML]{000000} \textbf{85.75}}} & \multicolumn{1}{c|}{\cellcolor[HTML]{FFFFFF}{\color[HTML]{000000} {\ul 85.89}}}    & \multicolumn{1}{c|}{\cellcolor[HTML]{FFFFFF}{\color[HTML]{000000} \textbf{86.76}}} & \multicolumn{1}{l|}{\cellcolor[HTML]{FFFFFF}}                   & \multicolumn{1}{c|}{\cellcolor[HTML]{FFFFFF}\textbf{94.75}}    & \multicolumn{1}{c|}{\cellcolor[HTML]{FFFFFF}\textbf{89.03}}    & \multicolumn{1}{c|}{\cellcolor[HTML]{FFFFFF}{ \textbf{93.49}}} & \multicolumn{1}{c|}{\cellcolor[HTML]{FFFFFF}\textbf{92.42}}       \\ \cline{1-1} \cline{3-6} \cline{8-11} 
\rowcolor[HTML]{FFFFFF} 
\multicolumn{1}{|l|}{\cellcolor[HTML]{FFFFFF}{\color[HTML]{000000} Ours (prune-rewind)}}              & \multicolumn{1}{l|}{\cellcolor[HTML]{FFFFFF}}                   & \multicolumn{1}{c|}{\cellcolor[HTML]{FFFFFF}{\color[HTML]{000000} { 87.63}}}    & \multicolumn{1}{c|}{\cellcolor[HTML]{FFFFFF}{\color[HTML]{000000} {\ul 83.13}}}    & \multicolumn{1}{c|}{\cellcolor[HTML]{FFFFFF}{\color[HTML]{000000} \textbf{85.99}}} & \multicolumn{1}{c|}{\cellcolor[HTML]{FFFFFF}{\color[HTML]{000000} {\ul 85.58}}}    & \multicolumn{1}{l|}{\cellcolor[HTML]{FFFFFF}}                   & \multicolumn{1}{c|}{\cellcolor[HTML]{FFFFFF}{\ul 94.68}}       & \multicolumn{1}{c|}{\cellcolor[HTML]{FFFFFF}{ 86.38}}       & \multicolumn{1}{c|}{\cellcolor[HTML]{FFFFFF}{\ul {93.18}}} & \multicolumn{1}{c|}{\cellcolor[HTML]{FFFFFF}{\ul 91.41}}          \\ \cline{1-1} \cline{3-6} \cline{8-11} 
\rowcolor[HTML]{EFEFEF} 
\multicolumn{1}{|c|}{\cellcolor[HTML]{EFEFEF}{\color[HTML]{000000} Oracle}}                             & \multicolumn{1}{l|}{\multirow{-9}{*}{\cellcolor[HTML]{FFFFFF}}} & \multicolumn{1}{c|}{\cellcolor[HTML]{EFEFEF}{\color[HTML]{000000} 92.34}}          & \multicolumn{1}{c|}{\cellcolor[HTML]{EFEFEF}{\color[HTML]{000000} 87.92}}          & \multicolumn{1}{c|}{\cellcolor[HTML]{EFEFEF}{\color[HTML]{000000} 88.80}}          & \multicolumn{1}{c|}{\cellcolor[HTML]{EFEFEF}{\color[HTML]{000000} 89.69}}          & \multicolumn{1}{l|}{\multirow{-9}{*}{\cellcolor[HTML]{FFFFFF}}} & \multicolumn{1}{c|}{\cellcolor[HTML]{EFEFEF}96.91}             & \multicolumn{1}{c|}{\cellcolor[HTML]{EFEFEF}92.13}             & \multicolumn{1}{c|}{\cellcolor[HTML]{EFEFEF}95.42}                & \multicolumn{1}{c|}{\cellcolor[HTML]{EFEFEF}94.82}                \\ \cline{1-1} \cline{3-6} \cline{8-11} 
\rowcolor[HTML]{FFFFFF} 
\multicolumn{11}{l}{\cellcolor[HTML]{FFFFFF}}                                                                                                                                                                                                                                                                                                                                                                                                                                                                                                                                                                                                                                                                                                                                                                                                                             \\ \cline{1-1} \cline{3-6} \cline{8-11} 
\rowcolor[HTML]{00009B} 
\multicolumn{1}{|c|}{\cellcolor[HTML]{00009B}{\color[HTML]{FFFFFF} }}                                   & \multicolumn{1}{l|}{\cellcolor[HTML]{FFFFFF}}                   & \multicolumn{4}{c|}{\cellcolor[HTML]{00009B}{\color[HTML]{FFFFFF} \textbf{Attribute: Arched Eyebrows}}}                                                                                                                                                                                                                                           & \multicolumn{1}{l|}{\cellcolor[HTML]{FFFFFF}}                   & \multicolumn{4}{c|}{\cellcolor[HTML]{00009B}{\color[HTML]{FFFFFF} \textbf{Attribute: Mouth Slightly Open}}}                                                                                                                                                             \\ \cline{3-6} \cline{8-11} 
\rowcolor[HTML]{DAE8FC} 
\multicolumn{1}{|c|}{\multirow{-2}{*}{\cellcolor[HTML]{00009B}{\color[HTML]{FFFFFF} \textbf{Methods}}}} & \multicolumn{1}{l|}{\cellcolor[HTML]{FFFFFF}}                   & \multicolumn{1}{c|}{\cellcolor[HTML]{DAE8FC}\textbf{Domain A}}                     & \multicolumn{1}{c|}{\cellcolor[HTML]{DAE8FC}\textbf{Domain B}}                     & \multicolumn{1}{c|}{\cellcolor[HTML]{DAE8FC}\textbf{Domain C}}                     & \multicolumn{1}{c|}{\cellcolor[HTML]{DAE8FC}\textbf{Average}}                      & \multicolumn{1}{l|}{\cellcolor[HTML]{FFFFFF}}                   & \multicolumn{1}{c|}{\cellcolor[HTML]{DAE8FC}\textbf{Domain A}} & \multicolumn{1}{c|}{\cellcolor[HTML]{DAE8FC}\textbf{Domain B}} & \multicolumn{1}{c|}{\cellcolor[HTML]{DAE8FC}\textbf{Domain C}}    & \multicolumn{1}{c|}{\cellcolor[HTML]{DAE8FC}\textbf{Average}}     \\ \cline{1-1} \cline{3-6} \cline{8-11} 
\rowcolor[HTML]{FFFFFF} 
\multicolumn{1}{|c|}{\cellcolor[HTML]{FFFFFF}Source only}                                               & \multicolumn{1}{l|}{\cellcolor[HTML]{FFFFFF}}                   & \multicolumn{1}{c|}{\cellcolor[HTML]{FFFFFF}72.94}                                 & \multicolumn{1}{c|}{\cellcolor[HTML]{FFFFFF}51.29}                                 & \multicolumn{1}{c|}{\cellcolor[HTML]{FFFFFF}56.71}                                 & \multicolumn{1}{c|}{\cellcolor[HTML]{FFFFFF}60.31}                                 & \multicolumn{1}{l|}{\cellcolor[HTML]{FFFFFF}}                   & \multicolumn{1}{c|}{\cellcolor[HTML]{FFFFFF}88.24}             & \multicolumn{1}{c|}{\cellcolor[HTML]{FFFFFF}80.36}             & \multicolumn{1}{c|}{\cellcolor[HTML]{FFFFFF}60.61}                & \multicolumn{1}{c|}{\cellcolor[HTML]{FFFFFF}76.40}                \\ \cline{1-1} \cline{3-6} \cline{8-11} 
\rowcolor[HTML]{FFFFFF} 
\multicolumn{1}{|c|}{\cellcolor[HTML]{FFFFFF}MEMO (AugMix)}                                             & \multicolumn{1}{l|}{\cellcolor[HTML]{FFFFFF}}                   & \multicolumn{1}{c|}{\cellcolor[HTML]{FFFFFF}72.72}                                 & \multicolumn{1}{c|}{\cellcolor[HTML]{FFFFFF}51.23}                                 & \multicolumn{1}{c|}{\cellcolor[HTML]{FFFFFF}56.38}                                 & \multicolumn{1}{c|}{\cellcolor[HTML]{FFFFFF}60.11}                                 & \multicolumn{1}{l|}{\cellcolor[HTML]{FFFFFF}}                   & \multicolumn{1}{c|}{\cellcolor[HTML]{FFFFFF}88.22}             & \multicolumn{1}{c|}{\cellcolor[HTML]{FFFFFF}80.30}             & \multicolumn{1}{c|}{\cellcolor[HTML]{FFFFFF}60.60}                & \multicolumn{1}{c|}{\cellcolor[HTML]{FFFFFF}76.37}                \\ \cline{1-1} \cline{3-6} \cline{8-11} 
\rowcolor[HTML]{FFFFFF} 
\multicolumn{1}{|c|}{\cellcolor[HTML]{FFFFFF}MEMO (RandConv)}                                           & \multicolumn{1}{l|}{\cellcolor[HTML]{FFFFFF}}                   & \multicolumn{1}{c|}{\cellcolor[HTML]{FFFFFF}72.66}                                 & \multicolumn{1}{c|}{\cellcolor[HTML]{FFFFFF}51.26}                                 & \multicolumn{1}{c|}{\cellcolor[HTML]{FFFFFF}56.44}                                 & \multicolumn{1}{c|}{\cellcolor[HTML]{FFFFFF}60.12}                                 & \multicolumn{1}{l|}{\cellcolor[HTML]{FFFFFF}}                   & \multicolumn{1}{c|}{\cellcolor[HTML]{FFFFFF}88.11}             & \multicolumn{1}{c|}{\cellcolor[HTML]{FFFFFF}80.27}             & \multicolumn{1}{c|}{\cellcolor[HTML]{FFFFFF}60.49}                & \multicolumn{1}{c|}{\cellcolor[HTML]{FFFFFF}76.29}                \\ \cline{1-1} \cline{3-6} \cline{8-11} 
\rowcolor[HTML]{FFFFFF} 
\multicolumn{1}{|c|}{\cellcolor[HTML]{FFFFFF}Ours (base)}                                               & \multicolumn{1}{l|}{\cellcolor[HTML]{FFFFFF}}                   & \multicolumn{1}{c|}{\cellcolor[HTML]{FFFFFF}76.39}                                 & \multicolumn{1}{c|}{\cellcolor[HTML]{FFFFFF}73.57}                                 & \multicolumn{1}{c|}{\cellcolor[HTML]{FFFFFF}{\ul {65.37}}}                  & \multicolumn{1}{c|}{\cellcolor[HTML]{FFFFFF}71.78}                                 & \multicolumn{1}{l|}{\cellcolor[HTML]{FFFFFF}}                   & \multicolumn{1}{c|}{\cellcolor[HTML]{FFFFFF}91.07}             & \multicolumn{1}{c|}{\cellcolor[HTML]{FFFFFF}82.49}             & \multicolumn{1}{c|}{\cellcolor[HTML]{FFFFFF}69.84}                & \multicolumn{1}{c|}{\cellcolor[HTML]{FFFFFF}81.13}                \\ \cline{1-1} \cline{3-6} \cline{8-11} 
\rowcolor[HTML]{FFFFFF} 
\multicolumn{1}{|c|}{\cellcolor[HTML]{FFFFFF}Ours (prune-zero)}                                       & \multicolumn{1}{l|}{\cellcolor[HTML]{FFFFFF}}                   & \multicolumn{1}{c|}{\cellcolor[HTML]{FFFFFF}\textbf{79.23}}                        & \multicolumn{1}{c|}{\cellcolor[HTML]{FFFFFF}\textbf{74.41}}                        & \multicolumn{1}{c|}{\cellcolor[HTML]{FFFFFF}63.57}                                 & \multicolumn{1}{c|}{\cellcolor[HTML]{FFFFFF}{\ul {72.40}}}                  & \multicolumn{1}{l|}{\cellcolor[HTML]{FFFFFF}}                   & \multicolumn{1}{c|}{\cellcolor[HTML]{FFFFFF}\textbf{92.36}}    & \multicolumn{1}{c|}{\cellcolor[HTML]{FFFFFF}\textbf{84.75}}    & \multicolumn{1}{c|}{\cellcolor[HTML]{FFFFFF}{\ul 73.31}}          & \multicolumn{1}{c|}{\cellcolor[HTML]{FFFFFF}{\ul {83.47}}} \\ \cline{1-1} \cline{3-6} \cline{8-11} 
\rowcolor[HTML]{FFFFFF} 
\multicolumn{1}{|l|}{\cellcolor[HTML]{FFFFFF}Ours (prune-rewind)}                                     & \multicolumn{1}{l|}{\cellcolor[HTML]{FFFFFF}}                   & \multicolumn{1}{c|}{\cellcolor[HTML]{FFFFFF}{\ul 78.26}}                           & \multicolumn{1}{c|}{\cellcolor[HTML]{FFFFFF}{\ul 73.91}}                           & \multicolumn{1}{c|}{\cellcolor[HTML]{FFFFFF}\textbf{69.68}}                        & \multicolumn{1}{c|}{\cellcolor[HTML]{FFFFFF}\textbf{73.95}}                        & \multicolumn{1}{l|}{\cellcolor[HTML]{FFFFFF}}                   & \multicolumn{1}{c|}{\cellcolor[HTML]{FFFFFF}{\ul 91.72}}       & \multicolumn{1}{c|}{\cellcolor[HTML]{FFFFFF}{\ul 83.11}}       & \multicolumn{1}{c|}{\cellcolor[HTML]{FFFFFF}\textbf{77.22}}       & \multicolumn{1}{c|}{\cellcolor[HTML]{FFFFFF}\textbf{84.02}}       \\ \cline{1-1} \cline{3-6} \cline{8-11} 
\rowcolor[HTML]{EFEFEF} 
\multicolumn{1}{|c|}{\cellcolor[HTML]{EFEFEF}Oracle}                                                    & \multicolumn{1}{l|}{\multirow{-9}{*}{\cellcolor[HTML]{FFFFFF}}} & \multicolumn{1}{c|}{\cellcolor[HTML]{EFEFEF}81.85}                                 & \multicolumn{1}{c|}{\cellcolor[HTML]{EFEFEF}72.94}                                 & \multicolumn{1}{c|}{\cellcolor[HTML]{EFEFEF}80.09}                                 & \multicolumn{1}{c|}{\cellcolor[HTML]{EFEFEF}78.29}                                 & \multicolumn{1}{l|}{\multirow{-9}{*}{\cellcolor[HTML]{FFFFFF}}} & \multicolumn{1}{c|}{\cellcolor[HTML]{EFEFEF}92.94}             & \multicolumn{1}{c|}{\cellcolor[HTML]{EFEFEF}88.39}             & \multicolumn{1}{c|}{\cellcolor[HTML]{EFEFEF}87.78}                & \multicolumn{1}{c|}{\cellcolor[HTML]{EFEFEF}89.70}                \\ \cline{1-1} \cline{3-6} \cline{8-11} 
\end{tabular}
}

\caption{\textbf{Domain-aware augmentation significantly improves generalization.} We report the single-shot SFDA performance (Accuracy \%) across different face attribute detection tasks and domain shifts. \alg~consistently improves upon MEMO while also performing competitively to the oracle. Through \textbf{bold} and \underline{underline} formatting, we denote the top two performing methods.}
\end{table*}

\noindent \textbf{Baselines.} In addition to the vanilla source-only baseline (no adaptation), we perform comparisons to the recent MEMO~\citep{zhang2021memo} technique - an online SFDA method which enforces prediction consistency between a image and its augmented variants. In particular, we implement MEMO with two popular augmentation strategies namely Augmix and RandConv~\citep{randconv}. Finally, we report the oracle performance \textit{i.e.}, NRC performance when all $12000$ unlabeled target data are available as opposed to our single-shot setting. 

\noindent \textbf{Findings.} From Table~\ref{tab:results}, it can be observed that, \alg~produces an average improvement of $\sim 10\%$ (across the three domain shifts) compared to the source-only baseline as well as the state-of-the-art MEMO. This improvement can be directly attributed to the efficacy of our generative augmentations, which can more effectively reflect the characteristics of the target domain than the pre-specified augmentations. While MEMO performs comparably to the source-only baseline at mild domain shifts (Domain A), it fairs poorly under severe shifts (Domain C). This clearly evidences the limitation of pixel-level corruptions used by MEMO in handling large domain shifts. Furthermore, with our approach, using the proposed activation pruning strategies leads to consistent improvements over the na\"ive sampling (base), due to the increased diversity in the curated target dataset. Finally, despite using only single-shot data, \alg~performs competitively to the oracle model obtained by using the entire target set (12K samples) for adaptation ($2\% - 6\%$ gaps on average).
\section{Conclusion}
In this paper, we explored the use of generative augmentations for test-time adaptation, when only a single-shot target is available. Through a combination of StyleGAN fine-tuning and novel sampling strategies, we were able to curate synthetic target datasets that effectively reflect the characteristics of any target domain. Our future work includes theoretically understanding the behavior of different pruning techniques and extending our approach beyond classifier adaptation.
\section*{Acknowledgements}
This work was performed under the auspices of the U.S. Department of Energy by the Lawrence Livermore National Laboratory under Contract No. DE-AC52-07NA27344. Supported by the LDRD Program under project 21-ERD-012.
\vskip 0.2in
\bibliography{refs}

\end{document}